%% file: iclr2027_conference.tex
\documentclass{article} 
\usepackage{iclr2027_conference,times}

\input{math_commands.tex}

\newcommand{\kernelname}[1]{\mathrm{#1}}

\usepackage{hyperref}
\usepackage{url}
\usepackage{booktabs}
\usepackage{multirow}
\usepackage{makecell}
\usepackage{graphicx}

\title{Compaction-Aware Reinforcement Learning for Ascend Triton Kernel Generation}

\author{%
\parbox{0.95\textwidth}{\centering
\bfseries
Chuansai Zhou\thanks{These authors contributed equally.}, Yanlin Liu\footnotemark[1],
Bingning Huang, Shidi Xi, Chao Huang, Boxing Chen, Mingming Zhu,
Guanghuan Fang, Ben Feng, Zongzuo Wang \\
\normalfont Huawei Technologies Ltd. \\
\scriptsize\texttt{%
\{zhouchuansai,liuyanlin5,huangbingning,boxing.chen\}@huawei.com,19149832005@bupt.edu.cn,\\
\{zhumingming1,fangguanghuan1,fengben,wangzongzuo\}@huawei.com,bruce.xi1@h-partners.com}}
}

\iclrfinalcopy
\begin{document}

\maketitle

\begin{abstract}
Automatically generating high-performance kernels with LLM-based agents
promises substantial gains in development efficiency, but two obstacles remain.
First, while increasing multi-turn interaction with execution environments is crucial for refining generation quality, continuous context accumulation rapidly exceeds the model's effective context window. Second, training long-horizon tasks via GRPO inherently incurs sparse rewards due to its reliance on terminal outcome evaluation, severely hindering step-level credit assignment; To address these obstacles, we propose ATRL (Ascend Triton Reinforcement Learning), a post-training method
that refines GRPO into a compaction-aware, segment-level objective for
long-horizon kernel generation. We also design a lightweight agent harness providing cascaded compilation, correctness, and profiling feedback, alongside an Ascend-specific knowledge base injected during agentic RL training and evaluation to compensate for the base model's missing platform expertise. Evaluations on NPUKernelBench and KernelBench, both
executed on Ascend hardware, show that ATRL consistently surpasses all
baselines, including GRPO, in both correctness and speedup.
\end{abstract}

\section{Introduction}

High-performance kernels are critically important in LLMs infrastructure, as they directly accelerate both the training and inference pipelines of LLMs. Recent advances in large language models (LLMs) have opened a promising path toward automated kernel generation. A growing body of work trains models to translate high-level PyTorch reference implementations into optimized kernels in domain-specific languages (DSLs) such as Triton~\cite{tillet2019triton}, whose tile-based programming model simplifies kernel development and substantially improves developer productivity~\cite{guo2026drtriton,liu2025drkernel}. To further improve generation quality, researchers have adopted an agentic paradigm: instead of producing a kernel through a single-round chat interaction, an LLM-based agent interacts with an execution environment across multiple turns--generating code, receiving compilation and correctness feedback, profiling performance, and iteratively refining its output~\cite{dai2026cudaagent,du2025adaexplore,lange2025robustkbench}. This multi-turn, tool-augmented workflow has proven substantially more effective than single-round generation, as it enables the model to learn from execution feedback and explore a broader optimization landscape.

Despite this progress, two fundamental challenges remain unresolved. First, the multi-turn agentic workflow inherently causes context to grow continuously: each turn appends tool results like view files, steadily inflating the input length. As recent workload characterizations have shown, this unbounded context growth not only degrades the quality of model inference, but also risks exceeding the effective context window, terminating the optimization loop prematurely~\cite{workload2025agentic,context2025escaping}. Existing kernel-generation agents~\cite{dai2026cudaagent,liu2025drkernel} either ignore this problem entirely or apply ad hoc truncation that discards potentially valuable debugging history, yielding no principled solution. Second, training long-horizon kernel generation with GRPO relies on a terminal outcome reward. Because credit is assigned only after the full trajectory finishes, intermediate edits receive little supervision, the learning signal remains sparse, and step-level credit assignment becomes unreliable. Standard GRPO therefore struggles to distinguish a useful repair from a later regression across many edit--verify turns.

To address these challenges, we propose \textbf{ATRL} (Ascend Triton Reinforcement Learning), a post-training method that refines GRPO for long-horizon kernel generation. ATRL decomposes a long agent trajectory into segments at verifiable boundaries and optimizes the policy with segment-level rewards built on three ideas: (i) \emph{structured trajectory compaction}, which condenses the accumulated history into a compact checkpoint once the context grows too long; (ii) \emph{mixed segment reward}, which combines the terminal task success with the verifiable progress measured at each segment's boundaries; and (iii) \emph{task-relative segment advantage}, which normalizes segment advantages across rollouts of the same task and weights them segment-wise in the loss. The compaction checkpoint itself participates in training, so the model learns both to preserve state and to continue from it. Throughout, a lightweight agent harness supplies compilation, correctness, and profiling feedback, and an on-demand knowledge base provides Ascend-specific optimization guidance.

\section{Related Work}
\subsection{LLM-Based Kernel Generation}
\textbf{Training-free agentic generation.} Several frameworks employ LLM agents that iteratively refine kernels through execution feedback without modifying model weights. AdaExplore~\cite{du2025adaexplore} introduces failure-driven adaptation, synthesizing recurring errors into reusable validity rules, combined with diversity-preserving tree search to escape local optima. The robust-kbench framework~\cite{lange2025robustkbench} proposes an evolutionary meta-generation procedure with LLM-based verifiers for correctness filtering, while also exposing exploitable benchmark loopholes. MKEvolve~\cite{yoo2026mkevolve} adopts a modular decomposition strategy, iteratively co-evolving sub-kernel decompositions and their implementations to improve interpretability and reduce token consumption. While effective at test time, these approaches do not improve the model's intrinsic generation ability and incur substantial inference cost per task.

\textbf{Training-based generation.} DRTriton~\cite{guo2026drtriton} proposes a synthetic data pipeline (CSP-DAG) with curriculum RL to train models for PyTorch-to-Triton conversion, achieving strong generalization despite training only on synthetic data. Dr.\,Kernel~\cite{liu2025drkernel} introduces KernelGYM, a distributed GPU environment with reward-hacking detection, and identifies a biased policy-gradient issue in GRPO, proposing Turn-level Reinforce-Leave-One-Out (TRLOO) for unbiased multi-turn advantage estimation. MusaCoder~\cite{wang2026musacoder} presents an end-to-end alignment pipeline for Moore Threads MUSA kernels. It proceeds from SFT through diversity-preserving RFT to two-stage RL, and introduces PrimeEcho, MirrorPop, and Buffered Dynamic Retry to stabilize training. Kernel-Smith~\cite{du2026kernelsmith} combines an evolutionary agent with post-training RL, converting long-horizon trajectories into step-centric supervision, and validates cross-backend transfer between NVIDIA Triton and MetaX MACA. CUDA-Agent~\cite{dai2026cudaagent} scales agentic RL with a skill-augmented environment and automated verification/profiling, achieving state-of-the-art results on KernelBench. Compiler-R1~\cite{pan2025compilerr1} applies RL to compiler pass-sequence auto-tuning, demonstrating that RL-trained LLMs can optimize IR-level metrics.

Despite this rapid progress, existing work is overwhelmingly centered on NVIDIA CUDA/Triton or, more recently, other GPU backends (MetaX MACA, Moore Threads MUSA). \emph{None of these methods address Ascend kernel generation}, where the Da Vinci architecture introduces fundamentally different compute and memory semantics. Moreover, most approaches treat the multi-turn execution environment as a black box for reward computation, without exploiting the transparency of intermediate verification steps for fine-grained credit assignment. Our work fills both gaps by targeting the Ascend Triton ecosystem and adopting a white-box RL formulation.

\subsection{Agentic RL}
\textbf{Agentic Reinforcement Learning.}
Agentic RL trains language models to improve through multi-turn interaction with tools and stateful environments, rather than optimizing isolated responses.
ArCHer establishes a hierarchical actor-critic formulation that combines turn-level value learning with token-level policy optimization~\citep{Zhou2024ArCHerTL}.
WebRL further addresses task scarcity, sparse feedback, and policy drift in web environments using a self-evolving curriculum and outcome-supervised rewards~\citep{Qi2024WebRLTL}.
For long-horizon digital agents, LOOP eliminates the learned critic through a leave-one-out baseline, enabling memory-efficient policy optimization directly in target environments~\citep{Chen2025ReinforcementLF}.
RAGEN instead studies the dynamics of multi-turn RL and identifies the Echo Trap, in which declining reward variance and policy entropy destabilize learning, motivating its stabilized StarPO-S algorithm~\citep{Wang2025RAGENUS}.

Recent studies address delayed credit assignment and unstable exploration through hierarchical or step-aware policy optimization.
GiGPO combines episode-level advantages with step-level comparisons among actions from equivalent states~\citep{Feng2025GroupinGroupPO}.
Agent Lightning decomposes arbitrary agent executions into trainable transitions through hierarchical credit assignment~\citep{Luo2025AgentLT}, while AgentGym-RL progressively increases interaction horizons to stabilize long-horizon training~\citep{xi2026agentgymrl}.
However, these methods are not readily applicable to Ascend kernel generation, where code states rarely recur and rewards are inherently sparse.
In contrast, our framework leverages compiler, correctness, and performance feedback to optimize long-horizon kernel-generation trajectories.

\section{Preliminary}
\subsection{Agentic Ascend Triton Kernel Generation}
\label{sec:prelim-kernel}

An Ascend Triton kernel-generation task starts from a reference PyTorch
implementation and a set of test inputs covering the required shapes and data
types. The goal is to produce a Triton kernel that matches the reference
outputs while running efficiently on the Ascend platform. Triton expresses the
kernel as a collection of parallel tile programs~\cite{tillet2019triton},
which the Ascend backend compiles into an executable kernel.

Producing a correct and efficient kernel usually requires several attempts.
The agent writes or modifies the kernel, invokes a unified evaluation tool,
and revises the implementation based on the structured report it
returns---compiler diagnostics, correctness results, and performance
measurements. This consolidated feedback makes the environment white-box,
allowing training to distinguish compilation failures, numerical errors, and
performance bottlenecks rather than relying only on final success; the full
verification cascade is detailed in Section~\ref{sec:agent-workflow} and
Appendix~\ref{app:verifier-details}.

Multi-turn optimization also creates an increasingly long interaction
history. Source-code snapshots, compiler logs, correctness reports, and
profiling results accumulate even after they are no longer useful. A long
history increases inference cost, distracts the model with stale information,
and may eventually exhaust its context window. Effective context management
is therefore necessary to sustain long-horizon kernel optimization.

\subsection{Group Relative Policy Optimization}
Group Relative Policy Optimization (GRPO)~\cite{grpo} is an online policy
optimization algorithm that avoids training a separate value model. For
each task $x$, it samples a group of $G$ trajectories
$\{\tau_i\}_{i=1}^{G}$ from the old policy. Let
$R_i=R(x,\tau_i)$ be the scalar reward of trajectory $\tau_i$. Its
group-relative advantage is
\begin{equation}
\widehat{A}_i
=
\frac{R_i-\mu_R}{\sigma_R},
\qquad
\mu_R=\frac{1}{G}\sum_{j=1}^{G}R_j,
\label{eq:grpo-advantage}
\end{equation}
where $\sigma_R$ is the standard deviation of the group rewards. This group-relative
baseline removes the critic required by conventional actor--critic methods
and directly compares candidate trajectories generated for the same task.

For each generated token, let $\rho_{i,t}$ be its importance ratio and define
the clipped ratio as
$\bar{\rho}_{i,t}=\kernelname{clip}(\rho_{i,t},1-\varepsilon,1+\varepsilon)$.
The GRPO objective is
\begin{equation}\mathcal{J}_{\mathrm{GRPO}}(\theta)=\mathbb{E}_{i,t}\!\left[\min\!\left(\rho_{i,t}\widehat{A}_i,\bar{\rho}_{i,t}\widehat{A}_i\right)-\beta D^{\mathrm{KL}}_{i,t}\right].\label{eq:grpo-objective}\end{equation}
where $D^{\mathrm{KL}}_{i,t}$ measures the divergence between the trainable
policy $\pi_\theta$ and a frozen reference policy $\pi_{\mathrm{ref}}$,
$\varepsilon$ is the clipping coefficient, and $\beta$ controls the KL
regularization strength. Following \cite{grpo}, the divergence is estimated
per token with the unbiased and nonnegative $k3$ estimator
\begin{equation}
D^{\mathrm{KL}}_{i,t}\approx
\frac{\pi_{\mathrm{ref}}(v_{i,t})}{\pi_\theta(v_{i,t})}
-\log\!\frac{\pi_{\mathrm{ref}}(v_{i,t})}{\pi_\theta(v_{i,t})}-1,
\label{eq:k3}
\end{equation}
where $v_{i,t}$ is the token generated at step $t$ of trajectory $\tau_i$ and
the token probabilities are conditioned on the trajectory prefix. The
expectation is taken over sampled tasks, trajectories, and policy-generated
tokens.

Recent variants relax the symmetric clipping interval of GRPO. In
particular, DAPO~\cite{yu2025dapo} clips the importance ratio asymmetrically,
$\kernelname{clip}(\rho_{i,t},1-\varepsilon_l,1+\varepsilon_h)$ with
$\varepsilon_h>\varepsilon_l$. This \emph{clip-higher} scheme enlarges the
upper bound so that low-probability tokens can still raise their probability
during optimization, counteracting the entropy collapse commonly observed in
long reinforcement learning runs and thereby encouraging exploration. ATRL
combines this asymmetric clipping with the KL-regularized GRPO objective;
Section~\ref{sec:compaction-rl} develops the resulting segment-level
formulation.

Standard outcome-supervised GRPO broadcasts the same trajectory-level
advantage $\widehat{A}_i$ to every generated token in $\tau_i$. This is
problematic for long-horizon kernel generation: unsuccessful early attempts,
useful diagnostic actions, and the final successful repair all receive the
same credit. Moreover, if most trajectories in a group fail compilation or
correctness checks, their rewards become identical or nearly identical,
producing little learning signal after group normalization. These
limitations motivate the trajectory decomposition and fine-grained reward
assignment introduced in Section~\ref{sec:method}.

\section{Method}
\label{sec:method}

We formulate Ascend Triton kernel generation as a long-horizon interaction between a language-model policy and an executable Ascend environment. Starting from a PyTorch reference implementation, the policy generates a kernel implementation, invokes verification tools, and revises the implementation using compiler, correctness, and profiling feedback. Two properties of this setting shape the design of ATRL. First, trajectories are long and grow without bound: a single kernel can require many edit--verify cycles, and the accumulated tool history eventually exceeds the model's context window. Second, the only naturally available scalar signal is the terminal outcome, which is too coarse to distinguish a useful repair from a later regression. ATRL addresses the first with a correctness-gated optimization workflow (Section~\ref{sec:agent-workflow}) and structured trajectory compaction (Section~\ref{sec:trajectory-compaction}), and the second with a compaction-aware RL objective that assigns credit to the resulting trajectory segments (Section~\ref{sec:compaction-rl}). Throughout, an on-demand knowledge base supplies Ascend-specific programming and optimization guidance (Section~\ref{sec:agent-workflow}).

\begin{figure}[t]
  \centering
  \includegraphics[width=0.86\linewidth]{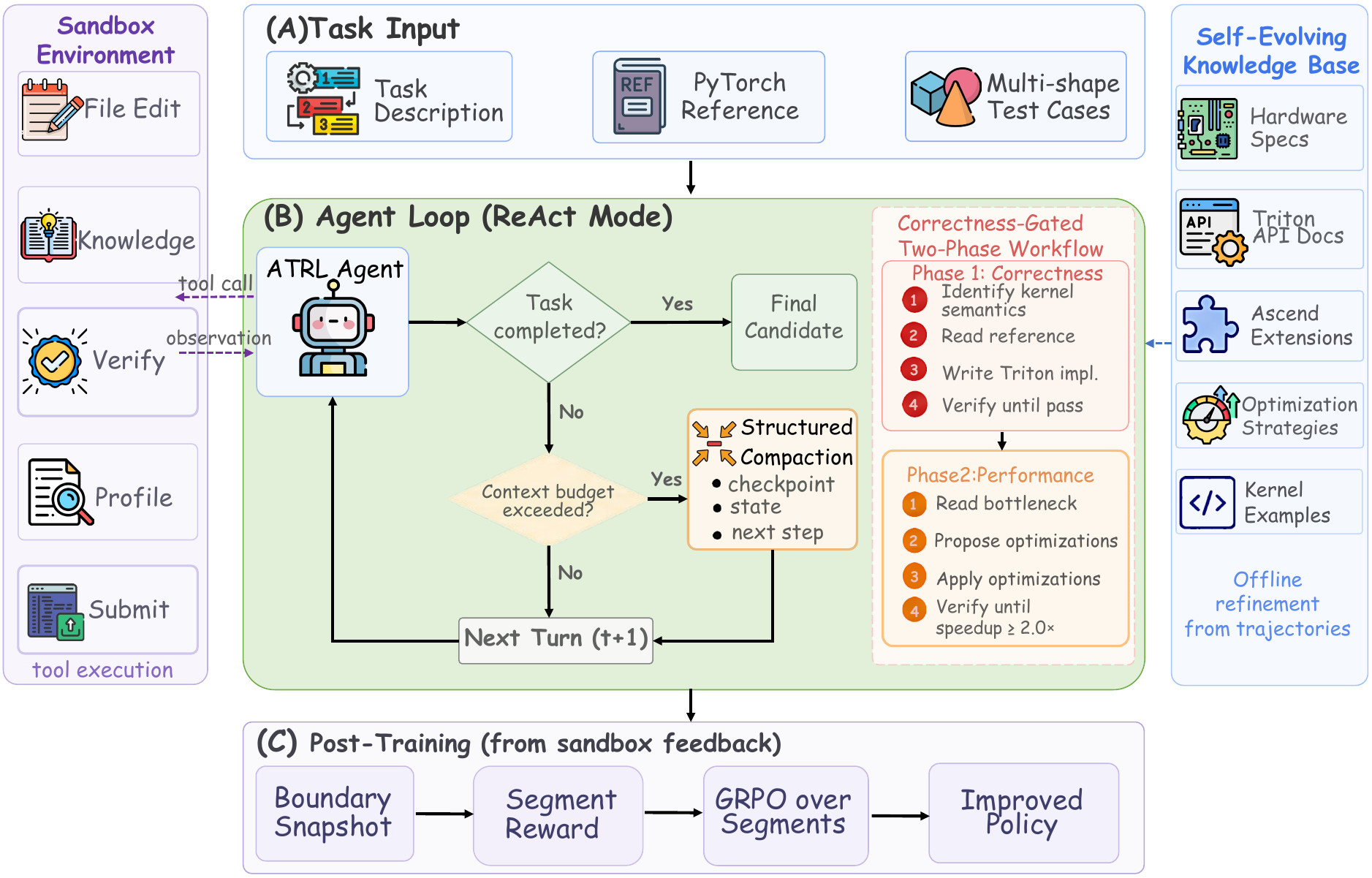}
  \caption{Method overview. The agent iteratively edits and verifies a Triton kernel in a sandbox. Long interactions are compacted into structured checkpoints, producing segments that receive both trajectory-level and incremental rewards.}
  \label{fig:method-overview}
\end{figure}

\subsection{Agentic Kernel Optimization}
\label{sec:agent-workflow}

\paragraph{Interaction loop.}
The agent operates in an isolated sandbox and alternates between language-model generation and tool execution. Its tools support file inspection and editing, knowledge retrieval, verification, and final submission; the full tool suite and sandbox setup are given in Appendix~\ref{app:harness-details}, and the corresponding prompt templates in Appendix~\ref{app:prompt-template}. Tool outputs are returned as observations in the next model context, allowing the policy to repair compilation errors, correct numerical mismatches, and optimize measured latency without requiring a complete kernel in a single response.

\paragraph{Correctness-gated optimization.}
We decouple correctness establishment from performance tuning by gating access to optimization guidance. During \emph{implementation}, the agent identifies the kernel semantics from the PyTorch reference, retrieves relevant API documentation, and iterates until the candidate passes verification. During \emph{optimization}, the agent uses profiling signals to select transformations such as tiling, multi-core partitioning, or improved memory access, while re-running correctness tests after every change. The gate prevents the policy from trading correctness for apparent speed. Near the turn limit, the harness may inject a brief reminder to create or verify an implementation; the reminder is also written to the rollout buffer so that inference and training observe the same sequence.

\paragraph{On-demand knowledge retrieval.}
The sandbox exposes a structured knowledge base of Ascend-specific guidance covering hardware constraints, Triton APIs and their Ascend extensions, example kernels, and optimization strategies, retrieved through a \emph{judge--retrieve--apply} chain that bounds token consumption. Before writing the kernel, the policy consults hardware constraints, examples, and API documentation to ground the initial implementation; during optimization, profiling signals guide the choice among transformations. Profiling reports diagnose symptoms---for example, low vector utilization or memory-bound execution---but do not prescribe an optimization, leaving the policy to connect each observation to an appropriate transformation. The retrieval interface and corpus maintenance are described in Appendix~\ref{app:knowledge-details}.

\paragraph{Cascaded verification.}
Each verification call executes a four-stage cascade---static analysis, compilation, numerical correctness, and hardware benchmarking---and stops at the first failed stage, so invalid candidates never reach the accelerator. The ordering is itself a safeguard: submissions that bypass the Triton kernel are rejected before any execution, and benchmarking runs only after every correctness case passes. Each stage emits a structured report that doubles as the environment observation. Stage definitions, precision criteria, and integrity safeguards are given in Appendix~\ref{app:verifier-details}.

\paragraph{Outcome reward.}
The verifier produces a scalar score that tracks progress from syntactic validity to execution efficiency:
\begin{equation}
\label{eq:reward}
R =
w_{\mathrm{ast}}\mathbf{1}_{\mathrm{ast}}
+ w_{\mathrm{cmp}}\mathbf{1}_{\mathrm{cmp}}
+ w_{\mathrm{cor}}p_{\mathrm{cor}}
+ w_{\mathrm{all}}\mathbf{1}_{\mathrm{all}}
+ w_{\mathrm{spd}}\min\!\left(\frac{s}{s^\star},1\right),
\end{equation}
where $\mathbf{1}_{\mathrm{ast}}$ and $\mathbf{1}_{\mathrm{cmp}}$ indicate that the candidate passes static analysis and compiles, respectively, $\mathbf{1}_{\mathrm{all}}$ indicates that all correctness cases pass, $p_{\mathrm{cor}}$ is the fraction of correctness cases passed, $s$ is the measured speedup, and $s^\star$ is the target speedup. The speed term is defined only when all correctness cases pass and is zero otherwise. We use
$(w_{\mathrm{ast}},w_{\mathrm{cmp}},w_{\mathrm{cor}},w_{\mathrm{all}},w_{\mathrm{spd}})
=(0.02,0.15,0.60,0.13,0.40)$ and $s^\star=2.0$. The maximum reward is $1.30$; a fully correct implementation with no speedup receives $0.90$, and fully correct rollouts therefore occupy $(0.90,1.30]$ as speedup increases to the target. Thus, an incorrect but fast implementation cannot receive performance credit.

\subsection{Structured Trajectory Compaction}
\label{sec:trajectory-compaction}

The iterative workflow above can produce trajectories that exceed the model's context window, making an explicit mechanism for preserving execution state necessary. Truncating old messages discards precisely the failures and constraints needed for subsequent repair. Instead, after a verification call and once context usage crosses a threshold, we ask the rollout model to replace the accumulated history with a structured checkpoint. The checkpoint records seven types of state: attempted approaches, discovered constraints, design decisions, errors and fixes, the current implementation, the latest verification result, and one immediate next action.

After compaction, the next context contains the unchanged system instruction and task, followed by the checkpoint and a short continuation instruction. The sandbox workspace and implementation files remain unchanged, so execution continues from the same environment state. Compaction is attempted only under conservative trigger conditions, and a checkpoint is accepted only if it is non-empty and shorter than the history it replaces.

Compaction partitions rollout $j$ into $I_j+1$ segments,
$\tau^j=\{\tau_1^j,\ldots,\tau_{I_j+1}^j\}$. At each boundary we store the code, verification status, and performance measurements. These snapshots allow us to measure what changed within a segment. We deliberately place the generated checkpoint in two positions: it is part of the preceding segment's response, where RL can supervise its generation, and part of the following segment's prompt, where it conditions continued execution. This dual role makes checkpoint quality itself optimizable while teaching the policy to preserve the state required for later kernel revisions. Trigger thresholds, prompt format, and failure-handling details appear in Appendix~\ref{app:compaction-details}.

\begin{figure}[t]
  \centering
  \includegraphics[width=0.86\linewidth]{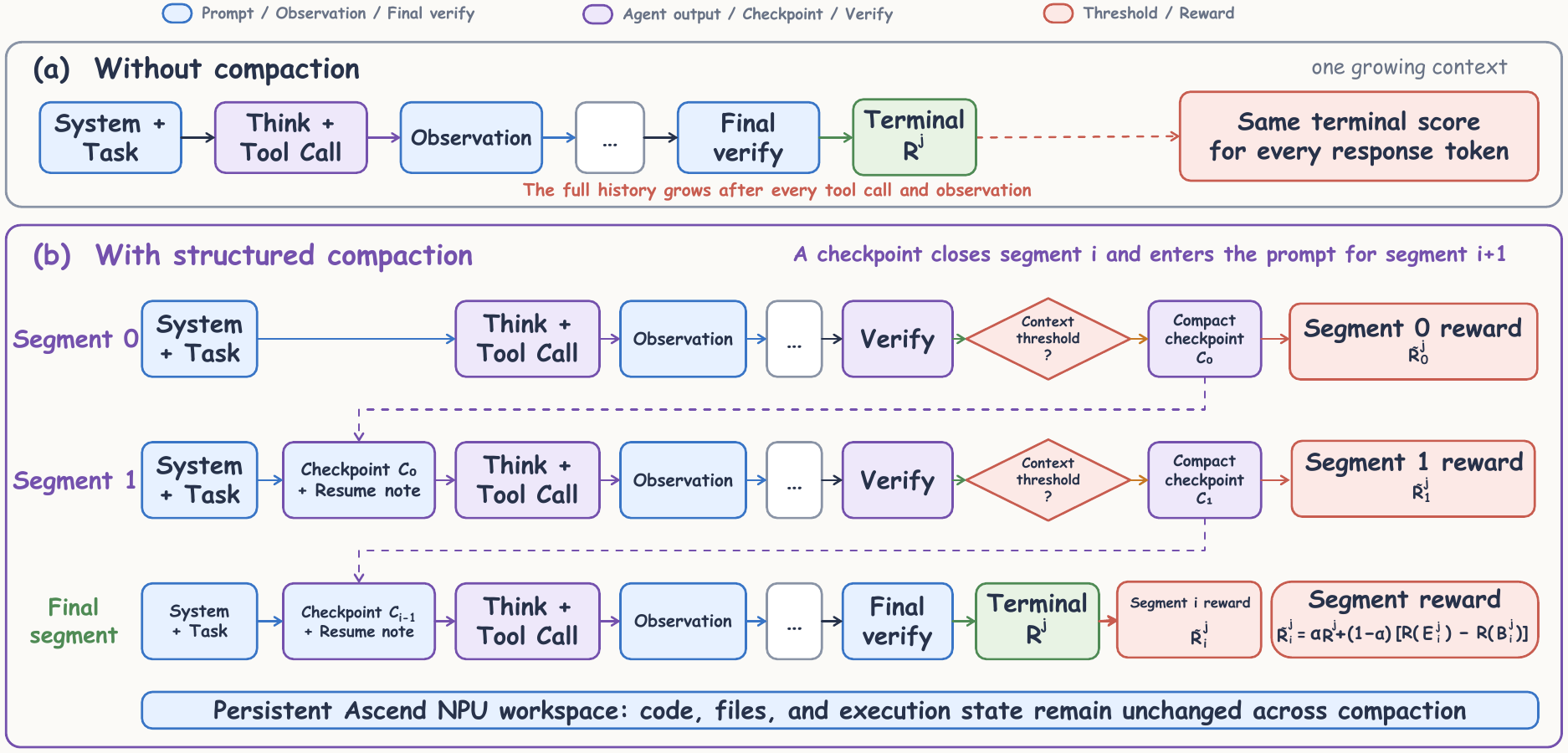}
  \caption{Compaction-aware segmentation. A checkpoint closes one training segment and initializes the prompt of the next; the sandbox workspace persists across the boundary.}
  \label{fig:trajectory-segments}
\end{figure}

\subsection{Compaction-Aware Reinforcement Learning}
\label{sec:compaction-rl}

Compaction changes the unit of training from one uninterrupted trajectory to several state-dependent segments. Standard trajectory-level GRPO assigns one terminal advantage to every generated token and cannot distinguish a useful repair from a later regression. We therefore combine global task success with the measured progress of each segment.

\paragraph{Segment reward.}
Let $R^j$ denote the score of the final verified implementation in rollout $j$. For segment $i$, let $B_i^j$ and $E_i^j$ denote its boundary states and define
\begin{equation}
\label{eq:local-reward}
r_i^j = R(E_i^j)-R(B_i^j),
\qquad
\widetilde{R}_i^j = \alpha R^j + (1-\alpha)r_i^j ,
\end{equation}
with $\alpha=0.5$ and $R(B_1^j)=0$. The first term preserves the end-to-end objective. The second attributes local progress: fixing compilation or correctness increases $r_i^j$, whereas breaking a previously correct kernel or reducing its speed makes $r_i^j$ negative. Both terms use Equation~\ref{eq:reward}, so they have the same scale. Setting $\alpha=1$ makes every segment receive the same terminal reward $R^j$, reducing the signal to the terminal outcome without per-segment progress; $\alpha=0$ retains only the local progress term.

\paragraph{Task-relative advantage.}
For one kernel task, we sample a group $g$ of rollouts and pool all of their segments. If
$N_g=\sum_{j\in g}(I_j+1)$, we compute
\begin{equation}
\label{eq:segment-advantage}
\mu_g=\frac{1}{N_g}\sum_{j\in g}\sum_i \widetilde{R}_i^j,
\quad
\sigma_g^2=\frac{1}{N_g}\sum_{j\in g}\sum_i
(\widetilde{R}_i^j-\mu_g)^2,
\quad
\widehat{A}_i^j=\frac{\widetilde{R}_i^j-\mu_g}{\sigma_g}.
\end{equation}
Normalizing within a task avoids comparing reward increments from kernels with different difficulty and optimization headroom. The segment advantage is broadcast only to response tokens in that segment.

\paragraph{Policy objective.}
We optimize a clipped policy-gradient objective:
\begin{equation}
\label{eq:objective}
\mathcal{J}(\theta)=
\mathbb{E}\!\left[
\frac{1}{G}\sum_{j=1}^{G}\frac{1}{I_j+1}\sum_{i=1}^{I_j+1}
\frac{1}{|a_i^j|}\sum_{t=1}^{|a_i^j|}
\left[
\min\!\left(
\rho_t\widehat{A}_i^j,\,
\kernelname{clip}(\rho_t,1-\varepsilon_l,1+\varepsilon_h)
\widehat{A}_i^j
\right)
-\beta D^{\mathrm{KL}}_t
\right]\right],
\end{equation}
where $\rho_t=\pi_\theta(v_t\mid x_t)/\pi_{\mathrm{old}}(v_t\mid x_t)$,
$\varepsilon_l$ and $\varepsilon_h$ are asymmetric clipping coefficients
following the clip-higher schedule of DAPO~\cite{yu2025dapo} (values in
Appendix~\ref{app:training-details}), and $D^{\mathrm{KL}}_t$ denotes the
per-token divergence from the frozen reference policy $\pi_{\mathrm{ref}}$,
estimated with the $k3$ estimator of Equation~\ref{eq:k3} and weighted by
$\beta$ as in GRPO~\cite{grpo}. Averaging first over tokens, then
segments, and finally rollouts prevents long responses or frequently
compacted trajectories from receiving disproportionate weight; we keep this
nested normalization rather than the flat token-level aggregation of DAPO
because it preserves segment-level credit assignment under trajectory
compaction.

\section{Experiments}
\label{sec:experiments}


\subsection{Setup}
\label{sec:exp-setup}

\textbf{Model and Datasets.}
We experiment with Qwen3.6-35B. Our training dataset is curated from the open-source cuda-llm%
\protect\footnote{\url{https://github.com/ByteDance-Seed/cudaLLM}} dataset, from which we select a subset of samples for RL training.

\textbf{Baselines.}
We compare: (1) untuned Qwen3.6-35B; (2) Qwen3.6-35B-SFT; (3) Dr.Kernel~\cite{liu2025drkernel}, using the open-source 14B%
\protect\footnote{\url{https://huggingface.co/hkust-nlp/drkernel-14b}} checkpoint (Drkernel-14B). We also train Qwen3-30B with this method, listed as Drkernel-30B; (4) GRPO, for which we train Qwen3.6-35B. The harness ablation additionally includes Qwen3.8-27B. Entries marked ``--'' are not evaluated.

\textbf{Benchmarks.}
We evaluate correctness and speedup on KernelBench~\citep{ouyang2024kernelbench}
and NPUKernelBench%
\protect\footnote{\url{https://github.com/Just-it/AscendOpGenAgent}.}. All performance measurements run on the Ascend platform.

\textbf{Evaluation Metrics.}
We assess results using two key metrics: (1) Pass@1, the percentage of tasks
where the agent generates a kernel that successfully compiles and passes functional correctness checks; We found that existing open-source evaluation standards are overly lenient, using atol=1e-2, rtol=1e-2, which permits some incorrect kernel implementations to pass precision validation. We instead adopt stricter thresholds: for fp32, atol=$10^{-3}$ and rtol=$2^{-13}$; for bf16, atol=$10^{-1}$ and rtol=$2^{-7}$. (2)Fast@1 and Fast@1.2, the fraction of generated kernels
whose measured speedup ratio over the reference reaches at least $1.0$ or
$1.2$.

\subsection{Main Results}
\label{sec:exp-results}

Tables~\ref{tab:kernelbench-exp} and \ref{tab:npukernelbench-exp} report the performance of ATRL and baselines on KernelBench and NPUKernelBench, respectively.

On KernelBench, ATRL achieves the best Pass@1 among reported entries on all three levels, reaching $73$ of $100$ on Level 1, $35$ of $100$ on Level 2, and $10$ of $50$ on Level 3. Level 3 is compared only with GRPO ($2$ of $50$ Pass@1); the other rows are not evaluated on this split. On Level 1, GRPO attains the highest Fast@1 and Fast@1.2, at $26\%$ and $23\%$, with ATRL at $23\%$ and $22\%$. On Level 2, ATRL attains the highest Fast@1 at $7\%$, and ties Drkernel-30B for the highest Fast@1.2 at $3\%$; SFT and GRPO are both at $2\%$ and $2\%$. On Level 3, Fast@1 and Fast@1.2 are $0\%$ for both GRPO and ATRL. The Pass@1 gain is consistent across levels. On Level 1 it does not raise Fast@1 or Fast@1.2 above GRPO. On Level 2 it coincides with higher Fast@1 ($7\%$ versus GRPO's $2\%$) and Fast@1.2 ($3\%$ versus $2\%$). On Level 3 both speedup rates stay at $0\%$.

On NPUKernelBench, ATRL achieves the best Pass@1 and Fast@1 on all four levels. The Pass@1 margin over GRPO is largest on Level~1 ($23$ versus $9$ of $33$) and Level~2 ($18$ versus $6$ of $31$). On Level 3, the most challenging split, ATRL obtains $9$ Pass@1, $4$ Fast@1, and $2$ Fast@1.2 out of $11$, compared with $3$, $1$, and $0$ for GRPO. This shows that ATRL is particularly effective in long-horizon, hardware-specific settings where correct kernel generation requires accumulating and reusing domain knowledge across multiple optimization turns.

\begin{table}[t]
\centering
\scriptsize
\setlength{\tabcolsep}{2.4pt}
\renewcommand{\arraystretch}{1.05}
\caption{Main Results on KernelBench. We report Pass@1 and Fast@1/1.2. Metrics are reported relative to torch eager baseline.
Each entry is the number of problems that meet the metric, out of 100, 100, and 50 problems on Levels 1--3. \textbf{Bold}
indicates the best result.}
\label{tab:kernelbench-exp}
\resizebox{\linewidth}{!}{%
\begin{tabular}{l *{3}{ccc}}
\toprule
\multirow{2}{*}{Method} &
\multicolumn{3}{c}{Level 1} &
\multicolumn{3}{c}{Level 2} &
\multicolumn{3}{c}{Level 3} \\
\cmidrule(lr){2-4} \cmidrule(lr){5-7}
\cmidrule(lr){8-10}
 & Pass@1 & Fast@1 & Fast@1.2
 & Pass@1 & Fast@1 & Fast@1.2
 & Pass@1 & Fast@1 & Fast@1.2 \\
\midrule
Drkernel-14B
 & 15 & -- & -- & 1 & -- & -- & -- & -- & -- \\
Drkernel-30B
 & 37 & 23 & 19 & 12 & 4 & \textbf{3} & -- & -- & -- \\
\midrule
Qwen3.6-35B
 & 20 & 8 & 7 & 1 & 0 & 0 & -- & -- & -- \\
Qwen3.6-35B-SFT
 & 33 & 12 & 11 & 4 & 2 & 2 & -- & -- & -- \\
GRPO
 & 66 & \textbf{26} & \textbf{23} & 19 & 2 & 2 & 2 & 0 & 0 \\
ATRL(ours)
 & \textbf{73} & 23 & 22 & \textbf{35} & \textbf{7} & \textbf{3} & \textbf{10} & 0 & 0 \\
\bottomrule
\end{tabular}%
}
\end{table}

\begin{table}[t]
\centering
\scriptsize
\setlength{\tabcolsep}{2.4pt}
\renewcommand{\arraystretch}{1.05}
\caption{Main Results on NPUKernelBench. We report Pass@1 and Fast@1/1.2. Metrics are reported relative to torch eager baseline.
Each entry is the number of problems that meet the metric, out of 20, 33, 31, and 11 problems on Levels 0--3. \textbf{Bold}
indicates the best result.}
\label{tab:npukernelbench-exp}
\resizebox{\linewidth}{!}{%
\begin{tabular}{l *{4}{ccc}}
\toprule
\multirow{2}{*}{Method} &
\multicolumn{3}{c}{L0} &
\multicolumn{3}{c}{L1} &
\multicolumn{3}{c}{L2} &
\multicolumn{3}{c}{L3} \\
\cmidrule(lr){2-4} \cmidrule(lr){5-7}
\cmidrule(lr){8-10} \cmidrule(lr){11-13}
 & Pass@1 & Fast@1 & Fast@1.2
 & Pass@1 & Fast@1 & Fast@1.2
 & Pass@1 & Fast@1 & Fast@1.2
 & Pass@1 & Fast@1 & Fast@1.2 \\
\midrule
Qwen3.6-35B
 & 12 & 7 & 5 & 7 & 2 & 1 & 3 & 3 & 3 & 3 & 3 & 2 \\
GRPO
& 11 & 5 & 4 & 9 & 1 & 0 & 6 & 1 & 0 & 3 & 1 & 0 \\
ATRL(ours)
& \textbf{18} & \textbf{12} & \textbf{8} & \textbf{23} & \textbf{4} & \textbf{3} & \textbf{18} & \textbf{8} & \textbf{8} & \textbf{9} & \textbf{4} & \textbf{2} \\
\bottomrule
\end{tabular}%
}
\end{table}


\subsection{Ablation Study}
\label{sec:exp-ablation-study}

We ablate the two components that address the obstacles stated in the introduction:
(i)~the lightweight agent harness of Section~\ref{sec:agent-workflow}, which turns the sandbox into a white-box environment, and
(ii)~the mixed segment reward of Section~\ref{sec:compaction-rl}, which replaces terminal GRPO credit assignment.
Table~\ref{tab:npukernelbench-exp-ablation} reports Pass@1 and Fast@1.

\begin{table}[t]
\centering
\scriptsize
\setlength{\tabcolsep}{2.4pt}
\renewcommand{\arraystretch}{1.05}
\caption{Ablation of the agent harness and the mixed segment reward.
Each entry is the number of problems that meet the metric. KernelBench Levels 1--3 contain 100, 100, and 50 problems; NPUKernelBench Levels 0--3 contain 20, 33, 31, and 11.}
\label{tab:npukernelbench-exp-ablation}
\resizebox{\linewidth}{!}{%
\begin{tabular}{l *{3}{cc}}
\toprule
\multicolumn{7}{c}{\textbf{(a) KernelBench}} \\
\midrule
\multirow{2}{*}{Method} &
\multicolumn{2}{c}{L1} &
\multicolumn{2}{c}{L2} &
\multicolumn{2}{c}{L3} \\
\cmidrule(lr){2-3} \cmidrule(lr){4-5}
\cmidrule(lr){6-7}
 & Pass@1 & Fast@1
 & Pass@1 & Fast@1
 & Pass@1 & Fast@1 \\
\midrule
Qwen3.6-35B$_{\scriptsize\text{w/o our harness}}$
 & 49 & 19 & 7 & 0 & 4 & \textbf{1} \\
Qwen3.6-35B$_{\scriptsize\text{w our harness}}$
 & 71 & 32 & 34 & 6 & 2 & 0 \\
\bottomrule
\end{tabular}%
}


\vspace{0.6em}

\resizebox{\linewidth}{!}{%
\begin{tabular}{l *{4}{cc}}
\toprule
\multicolumn{9}{c}{\textbf{(b) NPUKernelBench}} \\
\midrule
\multirow{2}{*}{Method} &
\multicolumn{2}{c}{L0} &
\multicolumn{2}{c}{L1} &
\multicolumn{2}{c}{L2} &
\multicolumn{2}{c}{L3} \\
\cmidrule(lr){2-3} \cmidrule(lr){4-5}
\cmidrule(lr){6-7} \cmidrule(lr){8-9}
 & Pass@1 & Fast@1
 & Pass@1 & Fast@1
 & Pass@1 & Fast@1
 & Pass@1 & Fast@1 \\
\midrule
Qwen3.6-35B$_{\scriptsize\text{w/o our harness}}$
 & 12 & 7 & 7 & 2 & 3 & 3 & 3 & 3 \\
Qwen3.6-35B$_{\scriptsize\text{w our harness}}$
 & 18 & 12 & 16 & 5 & 13 & 7 & 4 & 4 \\
\midrule
Qwen3.8-27B$_{\scriptsize\text{w/o our harness}}$
 & 15 & 10 & 16 & 9 & 14 & 10 & 5 & 1 \\
Qwen3.8-27B$_{\scriptsize\text{w our harness}}$
 & 19 & \textbf{13} & 15 & \textbf{10} & 17 & \textbf{11} & 6 & 3 \\
\bottomrule
\end{tabular}%
}
\end{table}


\paragraph{Effect of the agent harness.}
We compare the same multi-turn loop with versus without the harness: cascaded static analysis, compilation, numerical correctness, and benchmarking; correctness-gated access to profiling; and on-demand retrieval from the Ascend knowledge base through the judge--retrieve--apply chain.
On NPUKernelBench the largest relative gain is on Level~2 ($2$ to $13$ of $31$ Pass@1), where operators are both fused and hardware-specific; Qwen3.8-27B, a stronger starting point, still improves on Levels~0, 2, and 3, while Level~1 remains comparable.
On KernelBench, enabling the harness raises frozen Qwen3.6-35B from $49\%$ to $71\%$ Pass@1 on Level~1 and from $7\%$ to $34\%$ on Level~2, and raises Fast@1 on Level~1 from $19\%$ to $32\%$. On Level~3, Pass@1 falls from $4$ to $2$ of $50$ and Fast@1 from $1$ to $0$. This multi-turn comparison is separate from the Qwen3.6-35B row in Table~\ref{tab:kernelbench-exp}.
This supports the claim that the harness, including retrieval from the Ascend knowledge base, supplies checks and Triton-Ascend APIs that the base model does not apply on its own.




\section{Conclusion}
We introduced ATRL, a compaction-aware reinforcement learning method for long-horizon Ascend Triton kernel generation. ATRL preserves execution state through structured checkpoints and incorporates checkpoint generation into policy training. By combining terminal outcomes with verifiable progress at segment boundaries, it provides finer-grained credit assignment than outcome-supervised GRPO. A lightweight agent harness and an Ascend-specific knowledge base support iterative implementation, verification, and optimization. On KernelBench, ATRL improves Pass@1 over GRPO from $66\%$ to $73\%$ on Level~1 and from $19\%$ to $35\%$ on Level~2, with additional correctness gains on NPUKernelBench. Although performance gains vary across benchmark levels, these results support jointly designing context management and reward assignment for long-horizon kernel-generation agents. Future work will investigate stronger performance-oriented credit assignment and extend the approach to additional hardware backends.

\bibliography{iclr2027_conference}
\bibliographystyle{iclr2027_conference}

\appendix
\clearpage
\section{Agent Harness Details}
\label{app:harness-details}

\subsection{Tool Suite and Sandbox Execution}
The agent has file-viewing and incremental-editing tools, a knowledge catalog and document reader, a parameter-free verifier, and a submission tool. Tools are packaged as executable scripts, copied into the sandbox, and added to \texttt{PATH}.

\subsection{Termination and Final Scoring}
A rollout ends upon explicit submission, exhaustion of the turn or token budget, or sandbox timeout. If the code changes after the last verification, the harness performs one final verification before scoring. Whenever speedup improves, the harness saves the corresponding implementation and performance record as a historical-best snapshot. At each compaction boundary, the current verified state and the historical-best metadata are stored together, while segment rewards are computed from the current state.

\subsection{Knowledge Base}
\label{app:knowledge-details}
The knowledge base contains five modules:

\begin{itemize}
  \item hardware specifications, including core counts, on-chip memory, alignment, and transfer paths;
  \item a latency-optimization library with indexed strategies and operator-to-strategy mappings;
  \item Triton API documentation;
  \item Ascend-specific extensions for buffers, address spaces, and device functions; and
  \item complete kernel examples.
\end{itemize}

Documents are exposed at four levels of granularity: catalog overview, section, individual reference, and full text. Retrieval therefore proceeds as a \emph{judge--retrieve--apply} procedure: the policy first diagnoses what it needs, then retrieves only the targeted evidence, instead of loading whole documents into the context. The corpus is refined offline by aggregating frequent errors and performance bottlenecks from collected trajectories, then updating documentation, prompts, and tool diagnostics. These revisions modify the environment resources between training runs; the knowledge base itself is not updated by the policy during a rollout.

\section{Verifier Details}
\label{app:verifier-details}

\subsection{Pipeline Structure}
The verifier is a fail-fast cascade of four stages: static analysis, compilation, numerical correctness, and hardware benchmarking. Each stage runs only if its predecessor passed, and each writes a structured report that doubles as the observation returned to the agent. Compilation is exercised by importing and running the candidate in an isolated subprocess with a fixed time budget; import or launch failures are reported as compile errors. A stage passes only if every test case passes, and benchmarking additionally refuses to start unless the correctness report records a full pass, so no hardware time is spent on invalid candidates. Each stage terminates with a dedicated exit code (failure, correctness-gate refusal, missing baseline anchor, or tampered baseline), which the harness maps onto the indicator terms of Equation~\ref{eq:reward}. To keep observations compact, runtime errors are grouped by type, and each group retains one representative traceback after normalizing temporary paths and generated-source line numbers.

\subsection{Static Analysis}
The first stage parses the submitted source into an abstract syntax tree, without executing it, and rejects four classes of degenerate submissions: (i)~no Triton kernel at all, or kernels with empty bodies; (ii)~a kernel that is defined but never invoked from \texttt{ModelNew.forward}, where invocations are traced through same-file helper functions so that an unused kernel cannot be hidden behind an indirection; (iii)~a kernel that is invoked while part of the computation still runs through PyTorch, where \texttt{forward} may only call the kernel plus a whitelist of allocation and shape-manipulation utilities, and compute-heavy torch functions are rejected both in the forward call chain and at module scope; and (iv)~placeholder kernels, where a kernel counts as a placeholder if it uses no computational \texttt{tl.*} API and every value it stores is a variable whose only assignment is a \texttt{tl.load}, and a submission is rejected when all launched kernels are placeholders. The stage also blocks evasion techniques: dynamic attribute access into the \texttt{torch}/\texttt{nn} namespaces via \texttt{getattr}, direct calls to fused device operators, host-side evaluation of transcendental functions that precompute kernel input tables, and host--device synchronization in the launch path. Every rejection carries the regression class and a concrete repair suggestion that the agent can act on directly.

\subsection{Numerical Verification}
\textbf{Seeding protocol.} Inputs are generated, and both the reference and the candidate are instantiated, under identical seeds, so randomly initialized parameters agree bitwise.

\textbf{Hidden-state alignment.} For reference models whose output depends on internal state (randomly initialized weights, buffers, caches), a probe runs once before the first forward pass. Same-named parameters and buffers must match bitwise; the scan recurses into instance dictionaries and module-level caches, so weights hidden in plain containers are still found, and a transposed two-dimensional storage layout is accepted. The probe separates this failure mode---which no kernel repair can fix---from ordinary numerical mismatch and reports it before accuracy comparison.

\textbf{Comparison criteria.} Outputs must match in count and shape, and the positions (and for infinities, the signs) of non-finite values must match exactly. The remaining comparison dispatches on dtype: bool outputs require exact equality; integer outputs require exact equality for integer inputs but allow $|\Delta|\le 1$ for float inputs (quantization); float outputs must satisfy the three-part criterion in Table~\ref{tab:verifier-tols}; and non-computational operators, flagged per task, must produce bitwise-identical outputs of the same dtype, verified through integer reinterpretation so that even NaN payload differences are caught. Optional outputs that the reference returns as \texttt{None} must also be \texttt{None} in the candidate. Each test case runs in isolation with its own try/except, and failures are persisted with structured per-case metrics.

\begin{table}[h]
\centering
\caption{Float-precision thresholds by output dtype. A float output passes if (1)~every element satisfies $|\hat y - y| \le \mathrm{atol} + \mathrm{rtol}\,|y|$; (2)~the fraction of matched elements is at least $0.9$, where an element matches if $|y| < s_{\mathrm{thr}}$ implies $|\hat y - y| \le e_{\mathrm{max}}$, and otherwise $|\hat y - y|/(|y|+10^{-7}) \le r_{\mathrm{thr}}$; and (3)~the mean relative error $|\hat y - y|/(|y|+10^{-7})$ over all finite elements is below $r_{\mathrm{thr}}$. All quantities are computed in fp32.}
\label{tab:verifier-tols}
\small
\begin{tabular}{lccccc}
\toprule
dtype & atol & rtol & $s_{\mathrm{thr}}$ & $e_{\mathrm{max}}$ & $r_{\mathrm{thr}}$ \\
\midrule
fp32 & $10^{-3}$ & $2^{-13}$ & $2^{-14}$ & $2^{-30}$ & $2^{-13}$ \\
fp16 & $9\times10^{-2}$ & $2^{-10}$ & $2^{-11}$ & $2^{-16}$ & $2^{-10}$ \\
bf16 & $10^{-1}$ & $2^{-7}$ & $2^{-8}$ & $2^{-16}$ & $2^{-7}$ \\
\bottomrule
\end{tabular}
\end{table}

\subsection{Benchmarking Methodology}
Benchmarking measures each verified shape separately. Both models are instantiated under the seeding protocol above, and latency is collected with the profiler: after five warm-up iterations, five iterations are recorded, and only these enter the statistics. Per-shape latency is aggregated at the kernel level: durations are grouped by kernel name, per-kernel launch counts are recovered from row multiplicity so that multi-launch kernels are not underestimated, outlier samples are discarded, and the shape latency is the sum of per-kernel per-call averages. The per-shape speedup is the ratio of reference to candidate latency, and the overall speedup is the geometric mean over shapes with valid measurements. Profiling failures are retried up to three times per shape. Besides speedup, the stage reports peak memory and per-kernel utilization of the compute and data-movement pipes (vector, cube, and memory-transfer bandwidth); the agent reads these ratios to diagnose bottlenecks such as vector under-utilization or memory-bound execution and, guided by the knowledge base, to select transformations. Device frequency is locked during measurement.

\subsection{Integrity Safeguards}
When a task is frozen, the reference implementation and its test inputs are hashed; verification and benchmarking re-check these hashes and abort with a dedicated error if anything changed, so the agent cannot edit the reference, the inputs, or the scoring pipeline. Reference files are read-only in the sandbox. Fixed seeds, locked frequencies, and the per-stage subprocess timeout make verification outcomes reproducible.

\section{Context-Compaction Details}
\label{app:compaction-details}

\subsection{Trigger and Back-off}

The harness evaluates compaction after each agent turn. A compaction request is issued only when (i) compaction is enabled, (ii) at least six non-system/task messages are present, (iii) context usage exceeds a configured fraction of the model window, (iv) the most recent action was verification, and (v) the harness is not in a failure back-off period. We use exact usage reported by the inference service when available and otherwise estimate usage from message content plus role and JSON-serialization overhead. Training uses a lower threshold than evaluation to expose the policy to sufficient checkpoint-writing examples. Following a failed attempt, compaction is disabled for three interaction steps.

\subsection{Checkpoint Schema}

The compaction prompt requests the following sections in a fixed order:

\begin{enumerate}
  \item \textbf{Attempt history}: attempted approaches and their outcomes, merging repeated failures;
  \item \textbf{Learned constraints}: API, interface, AST, and hardware restrictions discovered so far;
  \item \textbf{Design decisions}: the rationale for the current implementation;
  \item \textbf{Errors and fixes}: observed failures and their resolutions;
  \item \textbf{Files and code}: relevant paths and the latest implementation;
  \item \textbf{Current state}: verification status, pass rate, speedup, and latest verifier output;
  \item \textbf{Next step}: one concrete continuation action.
\end{enumerate}

All sections must appear, with \texttt{none} used for empty fields. The model returns bullet-style text inside designated tags, may not call tools, and must merge valid information from any earlier checkpoint rather than stack summaries.

\subsection{Validation and Context Rebuilding}

We extract only text inside the designated tags and remove accidental code fences. A checkpoint is accepted only if it is non-empty and shorter than the replaced history. Compaction is attempted at most twice. Upon success, the context is rebuilt as

\[
[\text{system},\ \text{task},\ \text{checkpoint},\
\text{resume instruction}],
\]

where the resume instruction includes the checkpoint's next action and explicitly asks the model to continue from the existing workspace. On failure, the original conversation remains intact.

At each accepted boundary, the harness stores the messages and actions in the completed segment, together with token counts, compaction latency, current code, verification output, performance record, and historical-best speedup. It then opens a new rollout buffer. A rollout compacted $I$ times consequently yields $I+1$ rows in the training queue.

\section{Training Details}
\label{app:training-details}

\paragraph{Boundary rewards.}
Each segment's start state reuses the snapshot taken at the preceding compaction boundary, and segment-end scores are computed from the same verifier state used by the terminal reward. Therefore, improvements and regressions in AST validity, compilation, correctness, and speed are measured in a common unit.

\paragraph{Degenerate groups.}
All segments from rollouts of the same operator task participate in the statistics in Equation~\ref{eq:segment-advantage}. If the group contains only one segment, its standard deviation and advantage are set to zero because no relative comparison is available.

\paragraph{Loss masking.}
Environment observations, system messages, task instructions, and checkpoint text reused as a later prompt are excluded from the response loss. A checkpoint is optimized only when it is first generated as the response that closes the preceding segment. Token-level importance ratios are computed against the rollout policy, and the segment advantage is broadcast over the corresponding response mask.

\paragraph{Optimization hyperparameters.}
We use $\alpha=0.5$ for the mixed segment reward and asymmetric ratio clipping with $\varepsilon_l=0.20$ and $\varepsilon_h=0.28$. The KL penalty in Equation~\ref{eq:objective} applies the per-token $k3$ estimator of Equation~\ref{eq:k3} against the frozen reference policy, weighted by the coefficient $\beta$.


\section{Prompt Template}
\label{app:prompt-template}

\begin{figure*}[t]
    \centering
    \includegraphics[width=\textwidth]{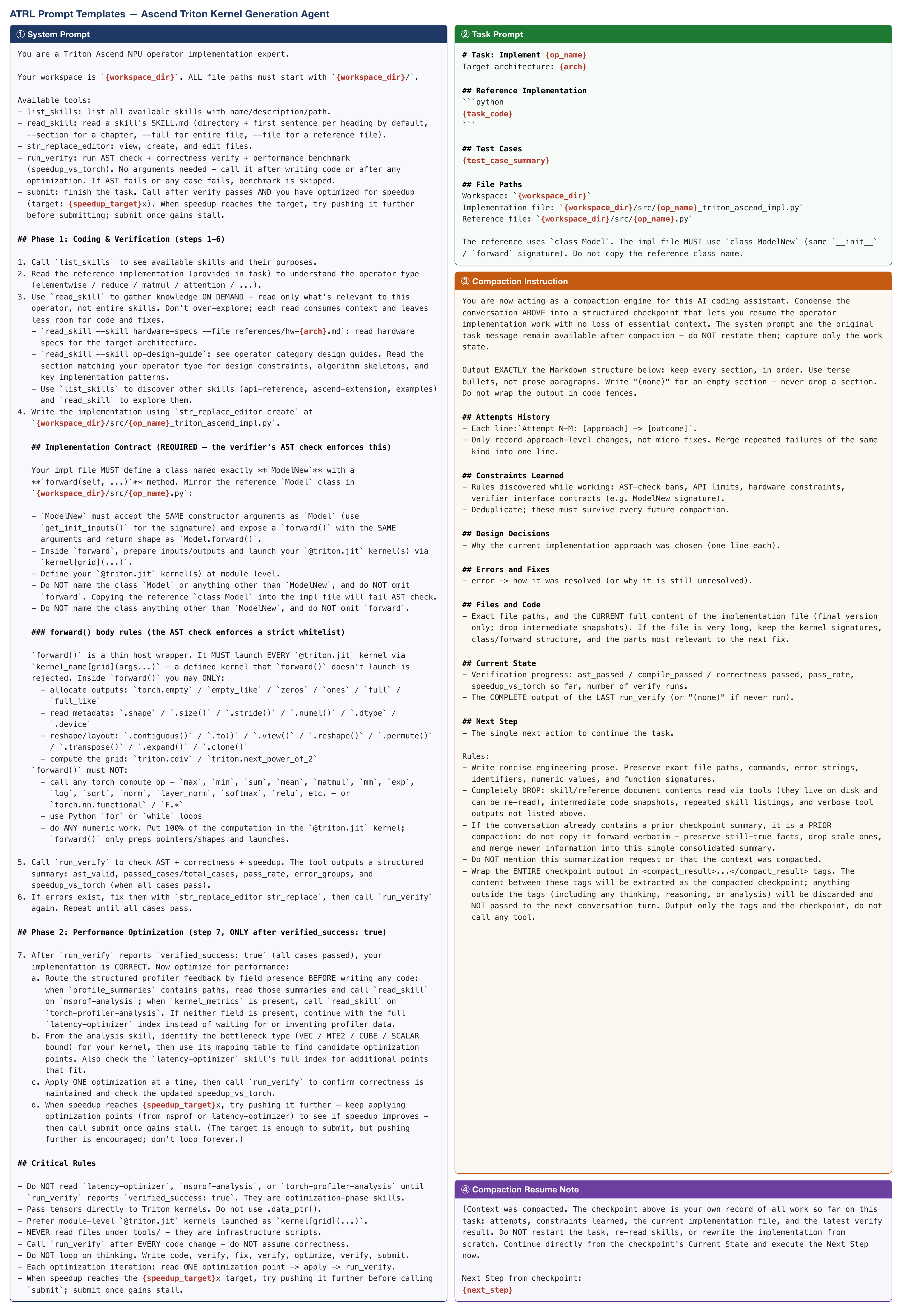}
    \caption{Prompt templates and context compaction instructions for the
    Ascend Triton Kernel Generation Agent.}
    \label{fig:atrl-prompt-templates}
\end{figure*}

Figure~\ref{fig:atrl-prompt-templates} presents the prompt templates used by the Ascend Triton Kernel Generation Agent.

\end{document}

%% file: math_commands.tex
\usepackage{amsmath,amsfonts,bm}

\def\eqref#1{equation~\ref{#1}}

\def\1{\bm{1}}

\DeclareMathAlphabet{\mathsfit}{\encodingdefault}{\sfdefault}{m}{sl}
\SetMathAlphabet{\mathsfit}{bold}{\encodingdefault}{\sfdefault}{bx}{n}

